\documentclass{article}

\usepackage[final, nonatbib]{neurips_2025}
\usepackage{booktabs} 
\usepackage{amsmath}
\usepackage{multirow}

\usepackage[utf8]{inputenc} 
\usepackage[T1]{fontenc}    
\usepackage{hyperref}       
\usepackage{url}            
\usepackage{booktabs}       
\usepackage{amsfonts}       
\usepackage{nicefrac}       
\usepackage{microtype}      
\usepackage{xcolor}         
\usepackage{graphicx}
\usepackage{wrapfig}

\usepackage{booktabs}
\usepackage{amssymb}
\usepackage{pifont}
\definecolor{mygreen}{RGB}{112, 180, 143}
\definecolor{myred}{RGB}{242, 128, 128}
\definecolor{citypink}{RGB}{227, 108, 194}
\definecolor{cityblue}{RGB}{128, 159, 225}

\newcommand{\Bench}{GEdit-Bench}
\newcommand{\cmark}{\textcolor{mygreen}{\ding{51}}}%
\newcommand{\xmark}{\textcolor{myred}{\ding{55}}}%

\usepackage{caption}
\usepackage{subcaption}
\usepackage{tikz}

\title{Step1X-Edit: A Practical Framework for General Image Editing}

\author{%
  Step1X-Image Team \\
  \texttt{StepFun} \\
  \href{https://github.com/stepfun-ai/Step1X-Edit}{\textcolor{citypink}{\textbf{\texttt{https://github.com/stepfun-ai/Step1X-Edit}}}}
}

\begin{document}

\maketitle

\begin{abstract}
In recent years, image editing technology has witnessed remarkable and rapid development. The recent unveiling of cutting-edge multimodal models such as GPT-4o and Gemini2 Flash has presented highly promising image editing capabilities. 
These models demonstrate an impressive aptitude for fulfilling a vast majority of user-driven editing requirements, marking a significant advancement in the field of image manipulation. 
However, there is still a large gap between the open-source algorithm with these closed-source models.
To this end, we introduce a state-of-the-art image editing model, Step1X-Edit, which aims to provide comparable performance against the closed-source models like GPT-4o and Gemini2 Flash. 
More specifically, we adopt the Multimodal LLM to process the reference image and the user's editing instruction. A latent embedding has been extracted and integrated with a diffusion image decoder to obtain the target image.
To train this model, we build a data generation pipeline covering 11 editing tasks to produce a high-quality dataset. For evaluation, we develop the \Bench, a novel benchmark rooted in real-world user instructions. 
Experimental results on \Bench~demonstrate that Step1X-Edit outperforms existing open-source baselines by a substantial margin and approaches the performance of leading proprietary models, thereby making significant contributions to the field of image editing.

\end{abstract}

\section{Introduction}
Image editing with natural language instructions has become an increasingly important task in vision-language research. It offers intuitive interaction for end users while posing unique technical challenges: understanding nuanced semantics, precisely localizing regions to edit, and preserving image fidelity. While diffusion models~\cite{podell2024sdxl,flux1dev,flux1schnell,esser2024scaling,peebles2023scalable} have dramatically improved image generation quality, the existing design by integrating text encoder, e.g., CLIP~\cite{clipRadford2021} and T5~\cite{2020t5}, with diffusion transformer often struggles with following editing instruction to maintain alignment between input image and edit instruction, especially when edit instructions are subtle or compositional.

Recent advances in proprietary multimodal foundation models, such as GPT-4o~\cite{gpt4o20250325}, Gemini2 Flash~\cite{gemini220250312}, and SeedEdit/Doubao~\cite{shi2024seededit}, have pushed the frontier of instruction-based image editing. These systems leverage large-scale vision-language modeling capabilities to perform high-fidelity edits across diverse scenarios. However, their closed nature limits reproducibility and transparency. In parallel, open-source efforts like OmniGen~\cite{xiao2024omnigen} and ACE++~\cite{mao2025ace++} aim to replicate similar capabilities but still fall short in terms of overall generalization, edit accuracy, and the quality of generated images.

In this work, we aim to narrow the performance gap between open-source and closed-source editing systems, while also pushing the boundary of practical and user-grounded editing evaluation. Although researchers have open-sourced editing datasets like AnyEdit~\cite{yu2024anyedit} and OmniEdit~\cite{wei2024omniedit}, we argue that the quality and diversity of these datasets are not good enough to obtain comparable performance against the close-source algorithms like GPT-4o. Thus, to target the image edit problem, we first try to build a large-scale high quality dataset for training. More specifically, we identify $11$ major editing task categories based on the commonly used editing instructions. Guided by this taxonomy, we develop a scalable and flexible data pipeline to generate over $1$ million high-quality training data. These image-instruction pairs encompass a broad spectrum of editing operations, including object manipulation, attribute modification, layout adjustment, and stylization, ensuring comprehensive coverage of real-world editing scenarios.

\begin{figure}[tbp]
    \centering
    \includegraphics[width=0.98\linewidth]{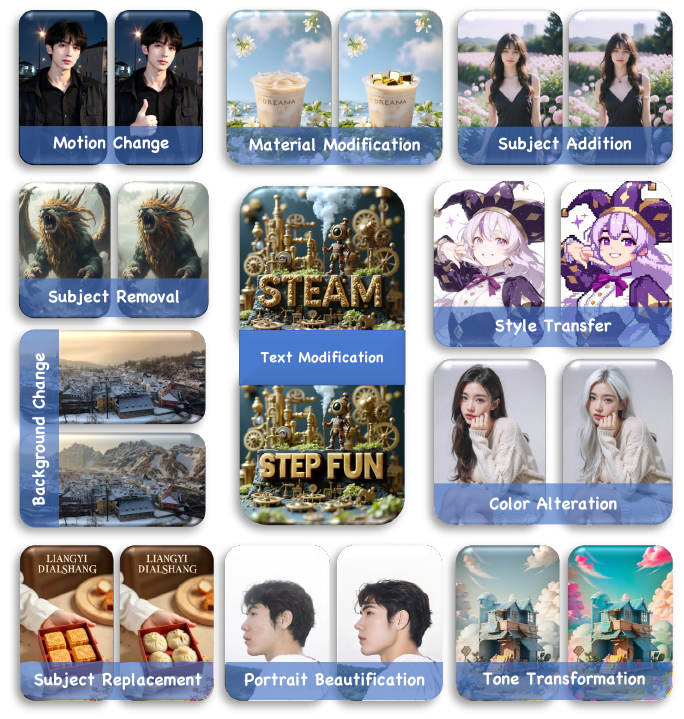}
    \caption{\textbf{Overview of Step1X-Edit}. Step1X-Edit is an open-source general editing model that achieves proprietary-level performance with comprehensive editing capabilities.}
    \label{fig:teaser}
\end{figure}

Building on this dataset, we propose,  \textbf{Step1X-Edit}, a unified image editing model that combines the strong semantic reasoning of Multimedia Large Language Model (MLLM), e.g., Qwen-VL~\cite{qwen2.5},  with a DiT-style diffusion architecture. The reference image and editing prompts can be processed by the MLLM to generate a target image latent condition which will be integrated with the diffusion model to obtain the output image. Our approach maintains a good balance between reference image reconstruction and editing prompt following. To train the model, we start from a text-to-image model to retain aesthetic quality and visual consistency, which can be easily replace by the existing text-to-image models like SD3~\cite{esser2024scaling}, FLUX\cite{flux1dev,flux1schnell,flux2024}, HiDream-I1~\cite{HiDream-I1} and  Flex~\cite{flex}.  To evaluate the existing editing models, we introduce a new benchmark named~\textbf{\Bench}. By carefully collecting the images and editing prompts, \Bench~ensures both real-world editing requirements and the diversity of the editing prompts. The experiments on~\Bench~validate that \textbf{Step1X-Edit} outperforms existing open-source baselines with a large-margin and approaches the performance of leading proprietary models, \textit{e.g.}, GPT-4o.

In summary, there will be three contributions of our work: 
\begin{itemize}
    \item We will open-source our \textbf{Step1X-Edit} model, to reduce the performance gap between open-source and closed-source image editing systems and boost further research in the field of image editing. 
    \item A data generation pipeline is designed to produce high-quality image editing data. It ensures that the dataset is diverse, representative, and of sufficient quality to support the development of effective image editing models. The availability of such a pipeline provides a valuable resource for researchers and developers working on similar projects.
    \item A new benchmark, named \textbf{\Bench}, grounded in real-world usages is developed to support more authentic and comprehensive evaluation. This benchmark, which is carefully curated to reflect actual user editing needs and a wide range of editing scenarios, enables more authentic and comprehensive evaluations of image editing models. 
\end{itemize}

\section{Related Work}

\subsection{Controllable Image Generation and Edit}

Autoregressive (AR) models have been actively studied for controllable image generation and editing by modeling images as sequences of discrete tokens. Works such as ControlAR~\cite{ControlAR}, ControlVAR~\cite{li2024controlvar}, and CAR~\cite{yao2024car} incorporate spatial and pixel-level guidance—such as edges, segmentation masks, and depth maps—into the decoding process, enabling localized and structured control. Extensions like Training-Free VAR~\cite{wang2025training}, M2M~\cite{shen2024many}, and Instruct-imagen~\cite{hu2024instruct} further improve editing flexibility and broaden application scenarios. UniFluid~\cite{fan2025unified} explores unified autoregressive generation and understanding with continuous visual tokens. However, due to reliance on discrete tokens and sequence length constraints, AR models often struggle to produce high-resolution and photorealistic results, especially in complex scenes.

Diffusion models have become the dominant approach for high-fidelity image synthesis, offering strong capabilities in photorealism, structural consistency, and diversity. Beginning with DDPM~\cite{ho2020denoising} and DDIM~\cite{Song2020DenoisingDI}, and further advanced by Latent Diffusion~\cite{Rombach2021HighResolutionIS,podell2024sdxl}, diffusion models operate in latent spaces for improved scalability. With the introduction of DiT architectures~\cite{peebles2023scalable}, diffusion models have made significant strides in generalization, image quality, and knowledge capacity, becoming the predominant architecture in modern image generation~\cite{stablediffusion3.5,flux1dev,flux1schnell}.
Based on the above text2image models, ControlNet~\cite{Zhang2023AddingCC}, and T2I-Adapter~\cite{Mou2023T2IAdapterLA} inject spatial or task-specific control into the generation process. BrushNet~\cite{ju2024brushnet}, PowerPaint~\cite{zhuang2024task}, and FLUX-Fill~\cite{flux1filldev2024} further improve inpainting quality and versatility. Despite these advances, diffusion models often rely on static prompts or fixed conditions and lack the capacity for multi-turn reasoning or flexible language alignment, limiting their application in open-ended editing scenarios.

These limitations have led to growing interest in unified image editing frameworks that combine the symbolic control of AR models with the generative fidelity of diffusion. Such models aim to tightly couple instruction understanding, spatial reasoning, and photorealistic synthesis within a single architecture, offering more flexible, general, and user-controllable editing capabilities. 

\subsection{Instruction-based Image Editing Models}

Instruction-based image editing models aim to bridge semantic instruction understanding and precise visual manipulation. Early approaches such as  InstructEdit~\cite{Wang2023InstructEditIA}, InstructPix2Pix~\cite{Brooks2022InstructPix2PixLT}, MagicBrush~\cite{zhang2023magicbrush}, and BrushEdit~\cite{Li2024BrushEditAI} adopt modular pipelines where MLLMs generate prompts, spatial cues, or synthetic instruction-image pairs to guide diffusion-based editing.

Recent works move toward tighter integration between instruction and generation. SmartEdit~\cite{Huang2023SmartEditEC}, X2I~\cite{Ma2025X2ISI}, RPG~\cite{Yang2024MasteringTD}, AnyEdit~\cite{yu2024anyedit}, and UltraEdit~\cite{zhao2024ultraedit} enhance multimodal interaction and instruction fidelity through improved model architectures, task-aware routing, and fine-grained editing capabilities. Meanwhile, unified generation and editing frameworks such as OmniGen~\cite{xiao2024omnigen}, ACE~\cite{han2024ace}, ACE++\cite{mao2025ace++}, and Lumina-OmniLV\cite{pu2025lumina} consolidate diverse visual tasks under a single architecture. Methods like Qwen2VL-Flux~\cite{erwold-2024-qwen2vl-flux}, DreamEngine~\cite{chen2025multimodalrepresentationalignmentimage} and MetaQueries~\cite{pan2025transfer} further explore efficient control integration and direct latent-level fusion between MLLMs and diffusion decoders. Moreover, Hidream-E1~\cite{HiDream-E1} incorporates instructions and edited image descriptions as inputs, enabling more detailed edit information.
Additionally, some researchers focus on more efficient methods to ensure editing performance while reducing training costs. For example, ICEdit~\cite{zhang2025ICEdit} employs LoRA-MoE hybrid tuning and identifies  better initial noise distributions, while SuperEdit~\cite{SuperEdit} uses higher-quality training data and introduces contrastive supervision signals.

More generally, models such as Gemini~\cite{gemini220250312} and GPT-4o~\cite{gpt4o20250325} demonstrate strong visual fluency through joint vision-language training, showing promising capabilities in understanding and generating consistent, context-aware images. Collectively, these developments reflect a shift from loosely coupled systems toward tightly integrated, instruction-driven editing frameworks.

However, existing approaches still face key limitations. Most methods are task-specific and lack general-purpose editability. They typically do not support incremental editing, fine-grained region correspondence, or instruction feedback refinement. Moreover, architectural coupling remains shallow in many designs, failing to unify instruction understanding and generation into a cohesive framework. These challenges motivate \textbf{Step1X-Edit}, which tightly integrates MLLM-based multimodal reasoning with diffusion-based controllable synthesis, enabling scalable, interactive, and instruction-faithful image editing across diverse editing goals.

\section{Step1X-Edit}
\subsection{Data Creation}
\subsubsection{Data Pipeline}

\begin{wrapfigure}{r}{0.5\textwidth}
  \centering
  \vspace{-10pt}  
  \includegraphics[width=0.5\textwidth]{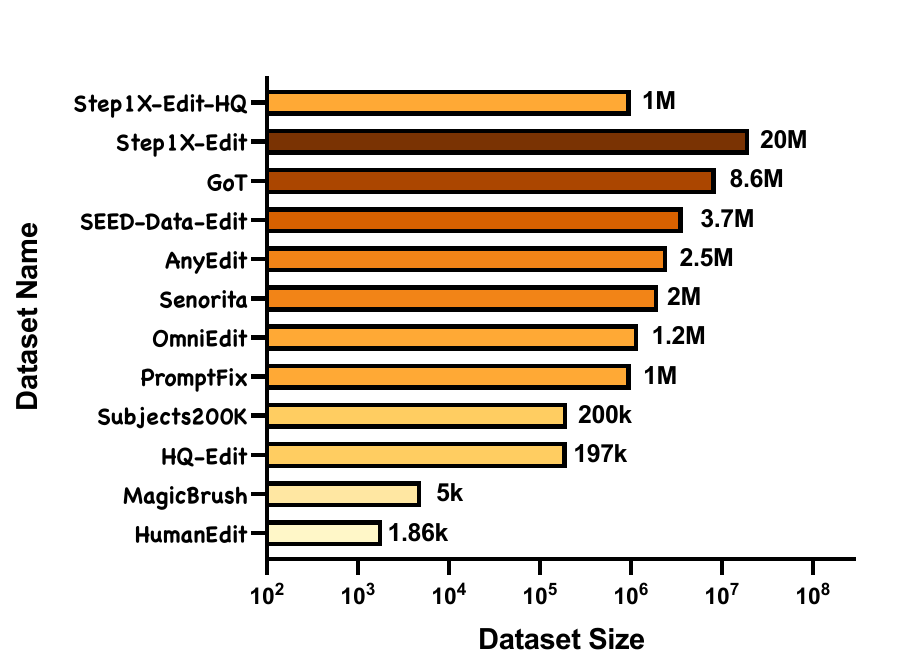}
  \caption{\small \textbf{Data Volume Comparison}.}
  \label{fig:stats}
  \vspace{-10pt}  
\end{wrapfigure}
\begin{figure}[htbp]
    \centering
    \includegraphics[width=0.92\linewidth]{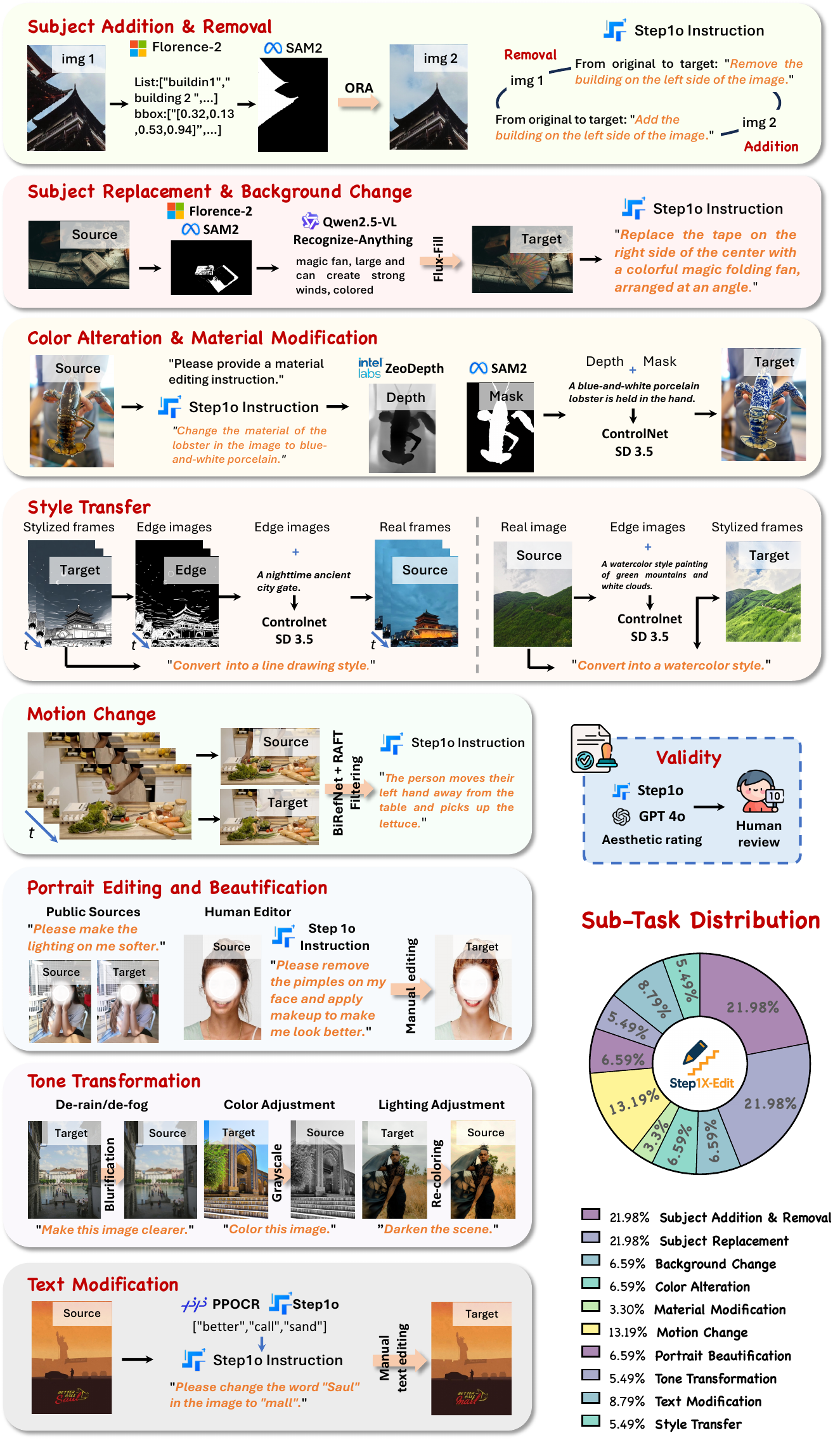}
    \caption{\textbf{Data Construction Pipeline and Sub-Task Distribution}.}
    \label{fig:data-pipeline}
\end{figure}

In the existing literature, current image editing datasets are constrained either by the scale or the quality of the collected data. To address this gap, this report endeavors to assemble a large-scale, high-quality dataset specifically tailored for image editing tasks.

We initiate the dataset collection process by web crawling a diverse set of image editing examples from the Internet. Through in-depth analysis of these examples, we systematically categorize the image editing problem into 11 distinct categories, which has been partly referenced by~\cite{yu2024anyedit,pan2025ice}. These categories are designed to comprehensively encompass the vast majority of image editing requirements in practice. An overview of these 11 categories, along with the detailed data collection pipeline, is illustrated in Fig.~\ref{fig:data-pipeline}.

To collect a large-scale high-quality triplets consisting of a source image, an editing instruction, and a target image,  we designed a sophisticated data pipeline, which enabled us to generate over $20$ million instruction-images triplets. Following rigorous filtering using both Multimodal LLMs, e.g. step-1o~\cite{step1o}, and human annotators, we retained more than $1$ million high-quality triplets. In Fig.~\ref{fig:stats}, we present a side-by-side comparison of all existing editing datasets~\cite{fang2025got,ge2024seed,zi2025senorita,wei2024omniedit,yu2024promptfix,tan2024ominicontrol,hui2024hq,zhang2023magicbrush,bai2024humanedit,yu2024anyedit}. Our Step1X-Edit dataset surpasses all others in scale. Even after a rigorous filtering process (with a retention ratio of 20:1), the \textbf{Step1X-Edit-HQ} subset remains on par with other datasets in terms of absolute magnitude. The full data collection pipeline for each subtask is outlined below.

\noindent\textbf{Subject Addition \& Removal}:
For subject-add and subject-remove tasks, we begin by annotating our proprietary dataset using Florence-2~\cite{xiao2024florence}, which supports diverse semantic granularities, spatial hierarchies, and annotation types such as object detection and classification. We then apply SAM-2~\cite{ravi2025sam} for segmentation and use ObjectRemovalAlpha~\cite{objectremovalalpha2025} to perform inpainting. Editing instructions are generated using a combination of Step-1o model~\cite{step1o} and GPT-4o, followed by manual review to ensure data validity.

\noindent\textbf{Subject Replacement \& Background Change}:
This category shares similar preprocessing steps with subject-add/remove, including Florence-2~\cite{xiao2024florence} annotation and SAM-2~\cite{ravi2025sam} segmentation. However, for these tasks, we utilize Qwen2.5-VL~\cite{qwen2.5} and the Recognize-Anything Model~\cite{zhang2023recognize} to identify target objects or keywords, followed by Flux-Fill~\cite{flux1filldev2024} for content-aware inpainting. The instructions are automatically generated by Step-1o and the triplets are human-verified.

\noindent\textbf{Color Alteration \& Material Modification}:
After detecting objects in the image, we employ Zeodepth~\cite{bhat2023zoedepth} for depth estimation to understand object geometry. Based on the identified target transformation (e.g., change of color or material), we use ControlNet~\cite{Zhang2023AddingCC} with diffusion model~\cite{stablediffusion3.5} to generate new images that preserve object identity while altering appearance attributes such as texture or color.

\noindent\textbf{Text Modification}:
For text-editing tasks, we differentiate between valid and invalid text edits. We use PPOCR~\cite{du2020ppocr}, which focuses on recognizing correct characters, alongside the Step-1o model to distinguish correct and incorrect regions of text. Based on this classification, we generate corresponding editing instructions. All outputs are finalized via human post-processing (e.g., manual retouching of text).

\noindent\textbf{Motion Change}:
To handle motion-related transformations, we leverage videos from Koala-36M~\cite{wang2024koala36mlargescalevideodataset}, extracting frame pairs as input. We use BiRefNet~\cite{zheng2024birefnet} and RAFT~\cite{teed2020raft} for foreground-background separation and optical flow estimation. Specifically, we compute the mean of the foreground flow norm and the norm of the background flow mean, ensuring robustness in selecting pairs where only the foreground exhibits motion. Finally, GPT-4o is used to annotate the change in motion between frames as editing instructions.

\noindent\textbf{Portrait Editing and Beautification}:
Data are collected and created by two major sources:
\textbf{(a)} Beautification pairs from public sources. Faces are detected and passed through Step-1o to assess layout and background consistency.
\textbf{(b)} Beautification of the human editor, we invite the human editor to conduct beatification on collected data. All data are manually validated.

\noindent\textbf{Style Transfer}:
We handle stylization in two directions depending on the target visual domain:
For styles such as Ghibli, ink painting, or 3D anime style, generating photorealistic images from stylized inputs yields better alignment. We extract edges from stylized images and generate realistic outputs using controlled diffusion model~\cite{Zhang2023AddingCC,stablediffusion3.5}.
Conversely, for styles like oil painting or pixel art, we begin with realistic images and generate stylized outputs using the same edge-to-image pipeline.

\noindent\textbf{Tone Transformation}:
This category focuses on global tonal adjustments, including color grading, dehazing, deraining, and seasonal transformations. These changes are largely driven by algorithmic tools and automated filters to simulate realistic environmental changes.

\subsubsection{Caption Strategy}

To obtain high-quality and fine-grained editing instruction–image pairs, we adopt the following annotation strategies:

    \noindent \textbf{Redundancy-Enhanced Annotation}:
Given the well-known limitations of Vision-Language Models (VLMs)—such as vague background descriptions and susceptibility to hallucinations—we employ a multi-round annotation strategy. Specifically, the annotation results from a previous round are fed into the next round as contextual input. This recursive refinement strengthens semantic consistency across annotations and significantly mitigates hallucination-related issues. Deterministic information is reinforced through repeated confirmations, ensuring higher reliability of the final annotation.

    \noindent \textbf{Stylized Annotation via Contextual Examples}:
During the captioning process, we provide annotators (or models) with a large set of style-aligned examples as contextual references. These examples guide the tone, structure, and granularity of the captions, ensuring a consistent and stylized annotation format throughout the dataset.

    \noindent \textbf{Bilingual Annotation (Chinese-English)}:
All our annotations are conducted bilingually, in both Chinese and English. This not only enhances accessibility and usability across different linguistic communities but also lays the groundwork for multilingual model training and evaluation.

\subsection{Our Method}
\begin{figure}[t!]
    \centering
    \includegraphics[width=0.98\linewidth]{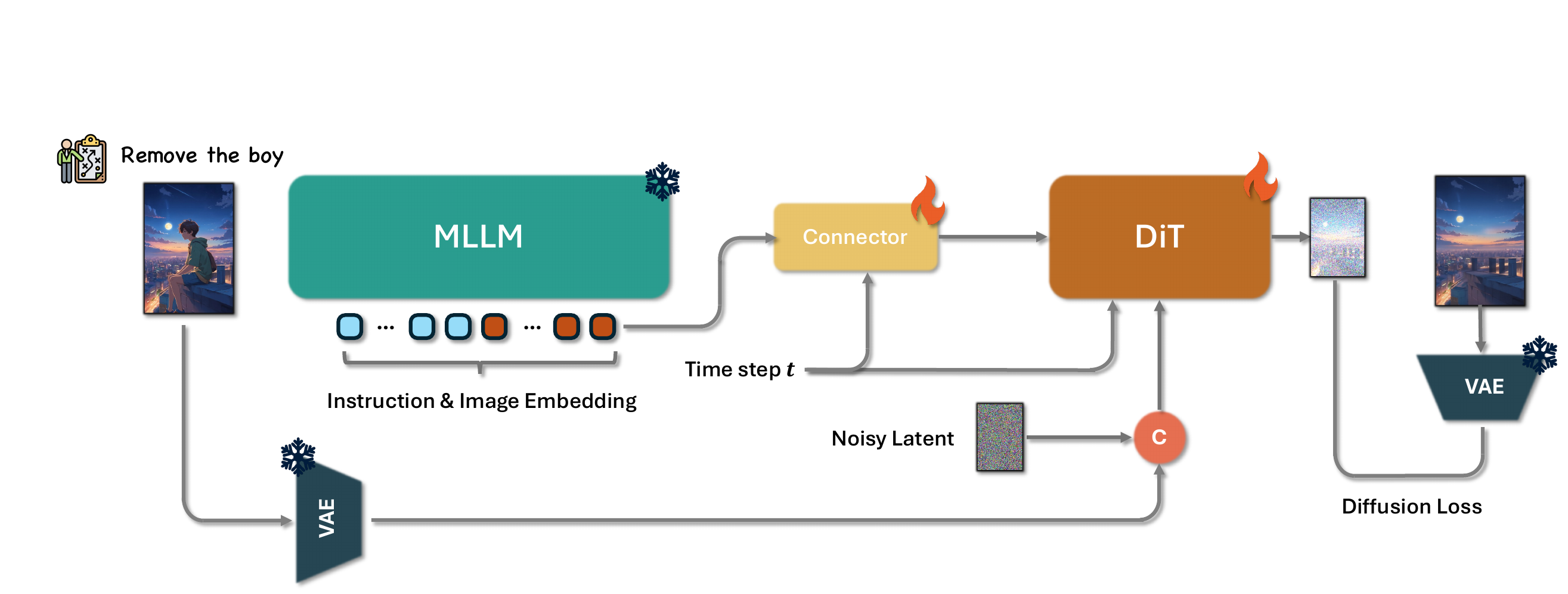}
    \caption{\textbf{Framework of Step1X-Edit}. Step1X-Edit leverages the image understanding capabilities of MLLMs to parse editing instructions and generate editing tokens, which are then decoded into images using a DiT-based network. }
    \label{fig:pipeline}
\end{figure}

As illustrated in Fig.~\ref{fig:pipeline}, our algorithm involves three key components: a Multimedia Large Language Model (MLLM), a connector module, and a Diffusion in Transformer (DiT)~\cite{peebles2023scalable}. The input editing instruction, accompanied by the reference image, is first introduced to the MLLM; e.g., Qwen-VL~\cite{qwen2.5} (hereafter abbreviated as Qwen). 
In conjunction with a system prefix, these inputs are jointly processed through a single forward pass of the MLLM, enabling the model to capture the semantic relationships between the instruction and the visual content. 
To isolate and emphasize the semantic elements relevant to the editing task, we selectively discard the token embeddings associated with the prefix. This filtering process retains only the token embeddings that directly align with the edit information, ensuring that subsequent processing focuses precisely on the editing requirements.

\begin{wrapfigure}{r}{0.33\textwidth}
  \centering
  \vspace{-10pt}  
  \includegraphics[width=0.33\textwidth]{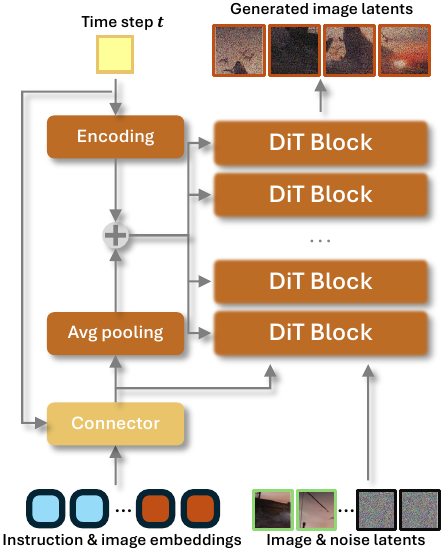}
  \caption{\small DiT module details.}
  \label{fig:DiT}
  \vspace{-10pt}  
\end{wrapfigure}
The extracted embeddings are then fed into a lightweight connector module, such as the token refiner~\cite{ma2024exploring,kong2024hunyuanvideo}. This module restructures the embeddings into a more compact multimodal feature representation, which is subsequently used as the multimodal embedding input to the downstream DiT network. Furthermore, we calculate the mean of valid embeddings from Qwen. This mean value is then projected through a linear layer, generating a global guidance vector. By doing so, the image editing network can leverage Qwen's enhanced semantic comprehension capabilities, enabling more accurate and context-aware editing operations.

To effectively train the Token Refiner and enable rich cross-modal conditioning, we carefully design our feature aggregation strategy. Compared to approaches such as FLUX-Fill~\cite{flux1filldev2024}, which uses channel concatenation, and methods like SeedEdit~\cite{shi2024seededit}, which introduce additional causal self-attention mechanisms, we follow OminiControl~\cite{tan2024ominicontrol} and adopt token concatenation to better balance responsiveness to editing instructions with the preservation of fine-grained image details.
During training, the reference image is encoded by a VAE encoder, and its latent features are linearly projected into reference image tokens. 
As illustrated in Fig.~\ref{fig:DiT}, the image tokens (highlighted in the green box) are concatenated with noise image tokens along the token length dimension to construct the final visual input. 

Recent research has also explored alternative approaches to enhance cross-model understanding by leveraging MLLMs. For instance, Qwen2VL-FLUX~\cite{erwold-2024-qwen2vl-flux} introduces a pioneering approach by replacing the traditional T5~\cite{2020t5} text encoder in DiT-based text-to-image models with MLLMs to enhance multi-modal understanding and generation. However, it still retains T5~\cite{2020t5} for text encoding, which limits its ability to perform comprehensive cross-modal reasoning. In contrast, approaches like DreamEngine~\cite{chen2025multimodalrepresentationalignmentimage} leverage Qwen to align image and text modalities, with their features serving as external conditional inputs for SD3.5~\cite{stablediffusion3.5} in the image synthesis. This method establishes a shared representation space that facilitates a more coherent process of generation and understanding. Nevertheless, in DreamEngine, the challenge remains in fully capturing the fine-grained details of reference images using only MLLM features.

Compared to these approaches, our model not only retains cross-modal understanding but also enhances the extraction of image details. 
By combining structured visual-language guidance, detailed visual condition, and strong pretrained backbones within a unified framework, our method significantly boosts the system's capability to perform high-fidelity, semantically aligned image edits across a diverse range of user instructions.
During training, we jointly optimize the connector and the downstream DiT using only the diffusion loss, following the rectified flow~\cite{esser2024scaling} formulation.
Moreover, our method ensures stable training without relying on the mask loss trick, which distinguishes it from OmniGen~\cite{xiao2024omnigen}.
The learning rate is fixed at $1e^{-5}$ to ensure a good trade-off between training stability and convergence speed.


\section{Benchmark and Evaluation}
\subsection{GEdit(Genuine Edit)-Bench}
To evaluate the performance of the image editing models, we collect a new benchmark called \textbf{GEdit(Genuine Edit)-Bench}. The main motivation of the benchmark is to collect the real-world user editing instances in order to evaluate how the existing editing algorithms can be suffice for the practical editing instructions. More specifically, we collect more than 1K user editing instances from the Internet, e.g., reddit, and manually split these editing instructions into the $11$ categories. To keep the diversity of the benchmark, we filter those editing instructions with similar purpose. Finally, we obtain 606 testing examples whose reference images are from the real-world cases which make it more genuine for the applications. 
Based on \textbf{\Bench}, we evaluate the existing open-source image editing algorithms like ACE++ and AnyEdit, as well as the closed-source algorithms like GPT-4o and Gemini2 Flash. 

\begin{wrapfigure}{r}{0.45\textwidth}
  \centering
  \vspace{-20pt}  
  \includegraphics[width=0.43\textwidth]{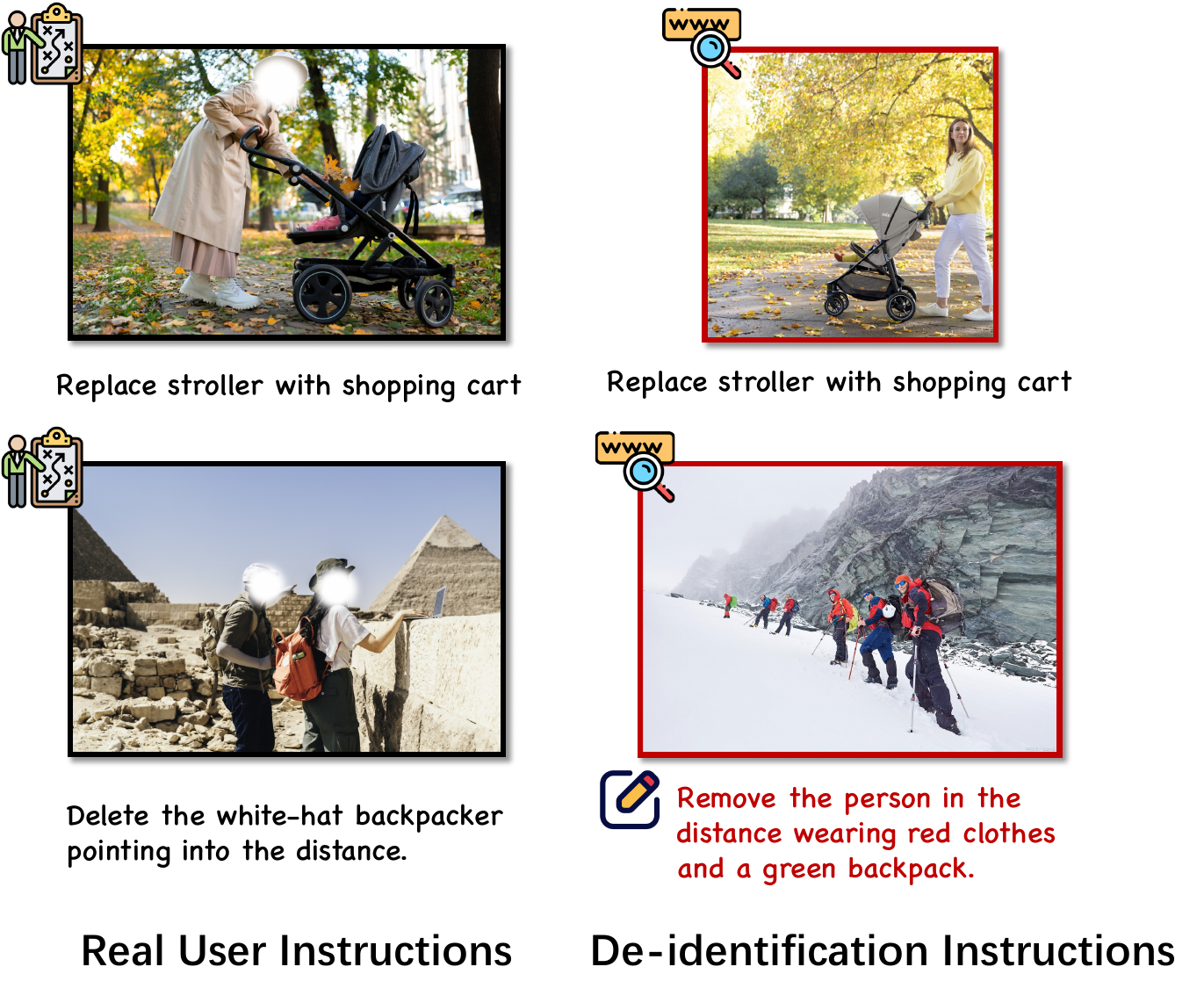}
  \caption{\textbf{De-Identification Process}.}
  \label{fig:half_right_image}
  \vspace{-10pt}  
\end{wrapfigure}


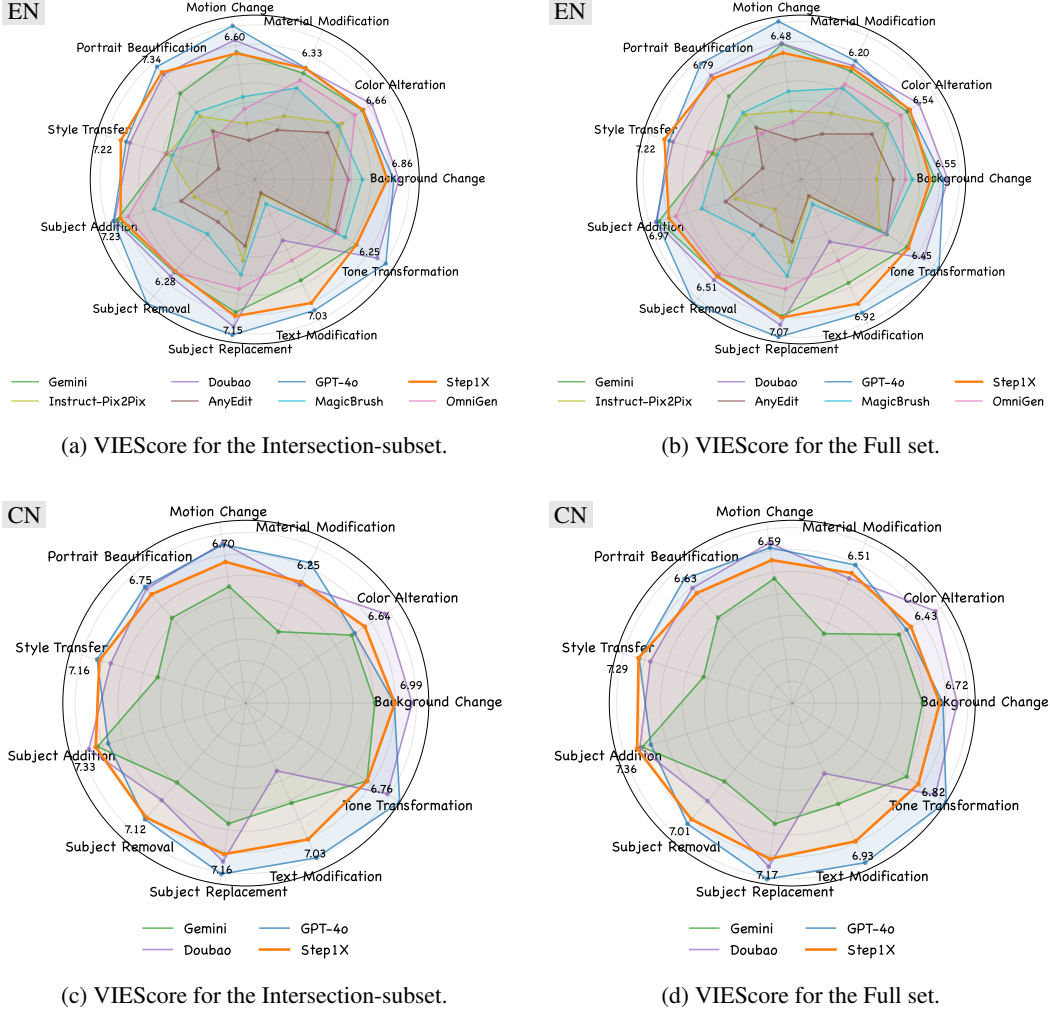
\begin{figure}[htbp]
    \centering

    \begin{subfigure}[b]{0.48\textwidth}
        \centering
        \begin{tikzpicture}
            \node[inner sep=0pt] (img) {\includegraphics[width=\linewidth]{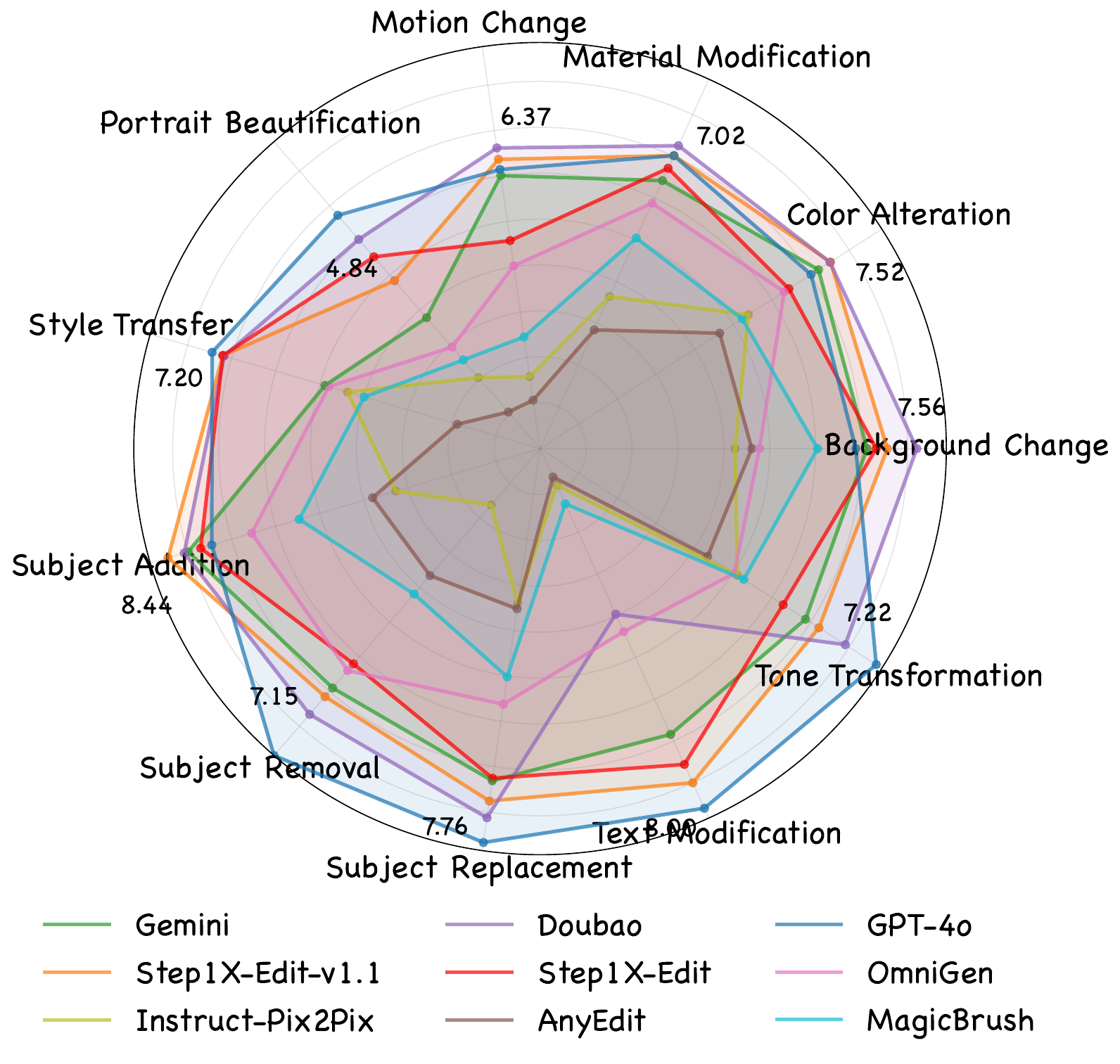}};
            \node[anchor=north west, fill=gray!20, text=black, font=\small, inner sep=2pt] at (img.north west) {EN};
        \end{tikzpicture}
        \caption{VIEScore for the Intersection-subset.}
    \end{subfigure}
    \hfill
    \begin{subfigure}[b]{0.48\textwidth}
        \centering
        \begin{tikzpicture}
            \node[inner sep=0pt] (img) {\includegraphics[width=\linewidth]{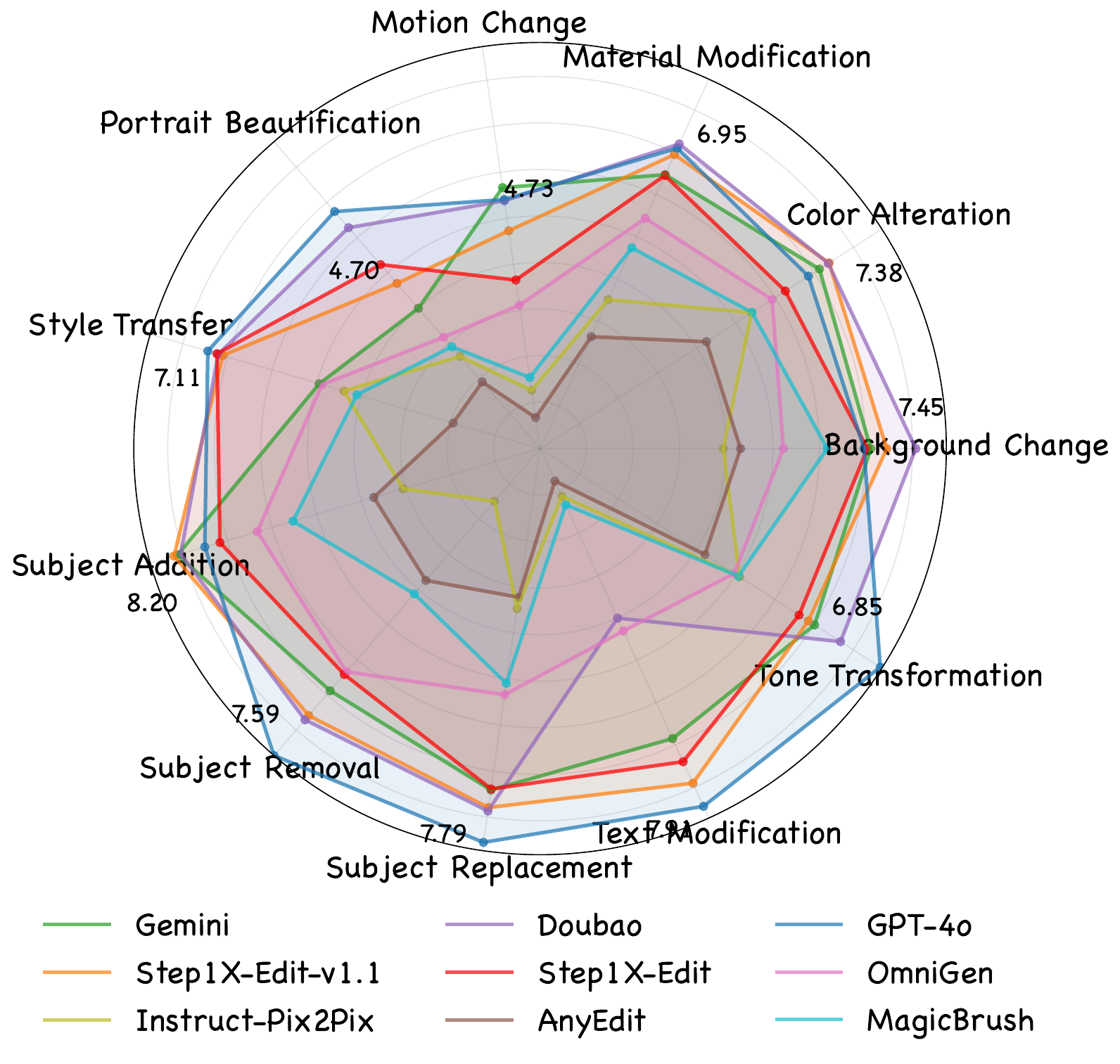}};
            \node[anchor=north west, fill=gray!20, text=black, font=\small, inner sep=2pt] at (img.north west) {EN};
        \end{tikzpicture}
        \caption{VIEScore for the Full set.}
    \end{subfigure}

    \vspace{0.5cm}

    \begin{subfigure}[b]{0.48\textwidth}
        \centering
        \begin{tikzpicture}
            \node[inner sep=0pt] (img) {\includegraphics[width=\linewidth]{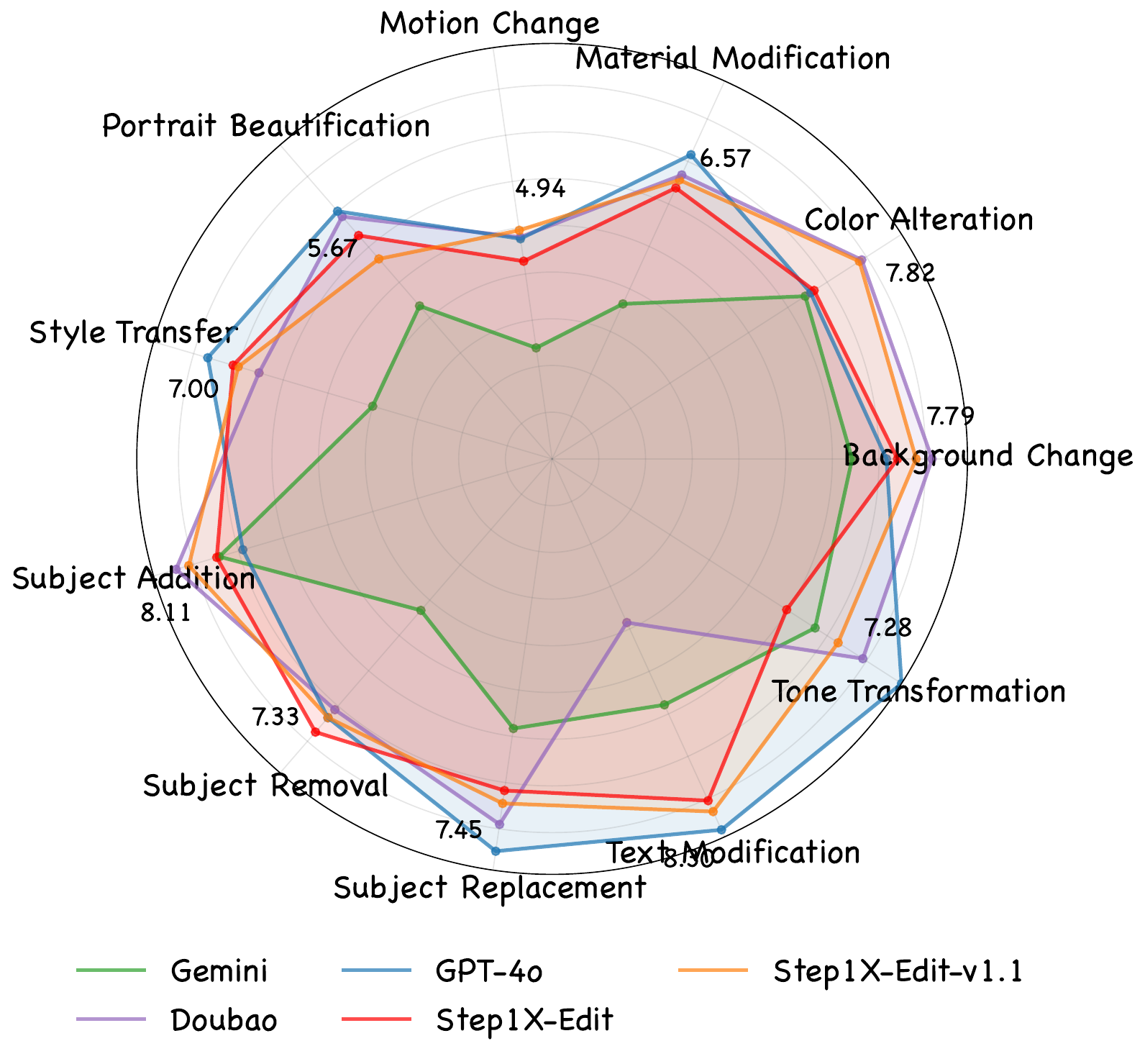}};
            \node[anchor=north west, fill=gray!20, text=black, font=\small, inner sep=2pt] at (img.north west) {CN};
        \end{tikzpicture}
        \caption{VIEScore for the Intersection-subset.}
    \end{subfigure}
    \hfill
    \begin{subfigure}[b]{0.48\textwidth}
        \centering
        \begin{tikzpicture}
            \node[inner sep=0pt] (img) {\includegraphics[width=\linewidth]{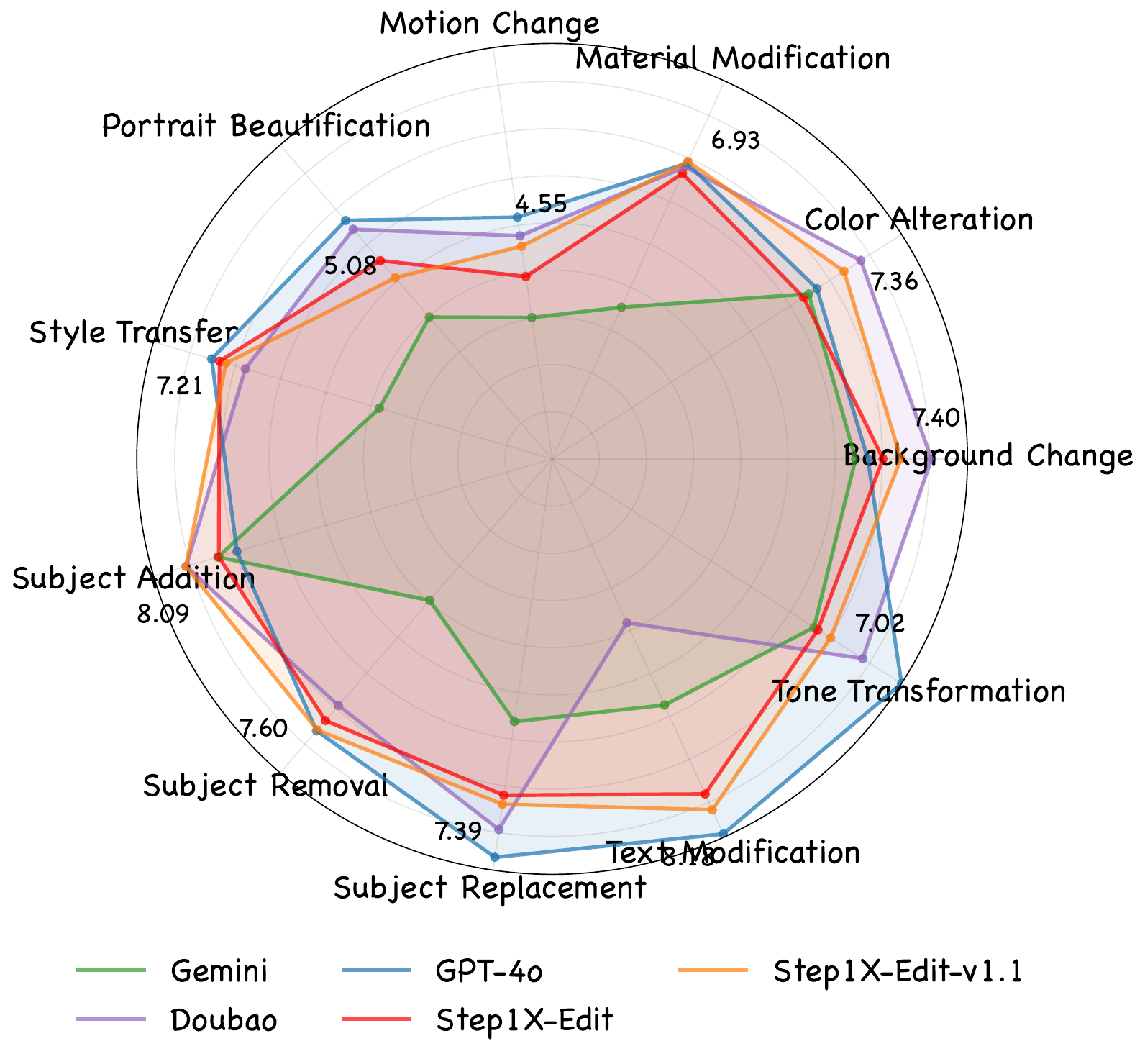}};
            \node[anchor=north west, fill=gray!20, text=black, font=\small, inner sep=2pt] at (img.north west) {CN};
        \end{tikzpicture}
        \caption{VIEScore for the Full set.}
    \end{subfigure}

    \caption{\textbf{VIEScore of Each Sub-task in \Bench. All the results are evaluated by GPT-4o}.}
    \label{fig:radar_gui_gpt}
\end{figure}

\begin{table}[t]
\centering
\resizebox{\textwidth}{!}{
\begin{tabular}{lcccccc}
\toprule
\textbf{Benchmarks} & \textbf{Size} & \textbf{Real Image} & \textbf{Genuine Instruction} & \textbf{Human Filtering} & \textbf{\#Sub-tasks} & \textbf{Public Availability} \\
\midrule
EditBench~\cite{wang2023imagen} & 240 & \cmark & \xmark & \xmark & 1 & \cmark \\
EmuEdit~\cite{sheynin2024emu} & 3,055 & \cmark & \xmark & \xmark & 7 & \cmark\\
HIVE~\cite{zhang2024hive} & 1,000 & \cmark & \xmark & \cmark & 1 & \cmark\\
HQ-Eidt~\cite{hui2024hq} & 1,640 & \xmark & \xmark & \xmark & 7 & \cmark\\
MagicBrush~\cite{zhang2023magicbrush} & 1,053 & \cmark & \xmark & \cmark & 7 & \cmark\\
AnyEdit~\cite{shen2024many} & 1,250 & \cmark & \xmark & \xmark & 25 & \cmark \\
ICE-Bench~\cite{pan2025ice} & 6,538 & \cmark & \xmark & \cmark & 31 & \xmark \\
\midrule
\textbf{\Bench (Ours)} & 606 & \cmark & \cmark & \cmark & 11 & \cmark\\
\bottomrule
\end{tabular}
}
\vspace{10pt}
\caption{\textbf{Key Attributes of Open-source Edit Benchmarks}. The reliance of existing open-source benchmarks on synthetic user inputs and minimal human involvement highlights the necessity of our proposed \Bench.
}
\label{tab:dataset_comparison}
\end{table}

To safeguard privacy, a comprehensive de-identification protocol was meticulously implemented for all user-uploaded images prior to their utilization within the benchmarking framework as shown in Fig.~\ref{fig:half_right_image}. For each individual original image, a multi-faceted reverse image search strategy was employed, spanning across multiple public search engines. This process aimed to identify publicly accessible alternative images that demonstrated both visual similarity and semantic consistency with the original one, thereby aligning seamlessly with the corresponding editing instructions. In instances where public image alternatives could not be procured through this search methodology, a systematic approach to modifying the editing instructions was adopted. These modifications were carefully calibrated to maintain the highest degree of fidelity between the anonymized image-instruction example and the original user intents. This approach not only ensures the ethical integrity of the benchmark dataset but also preserves the essential characteristics required for accurate and meaningful evaluation of image editing models.

\begin{table*}[ht]
\centering
\resizebox{\textwidth}{!}{%
\begin{tabular}{l|ccc|ccc|ccc|ccc}
\toprule
\multirow{3}{*}{\textbf{Model}} 
& \multicolumn{6}{c|}{\textbf{\Bench-EN (Intersection subset)} $\uparrow$} 
& \multicolumn{6}{c}{\textbf{\Bench-EN (Full set)} $\uparrow$} \\
\cmidrule(lr){2-7} \cmidrule(lr){8-13}
& \textbf{G\_SC} & \textbf{G\_PQ} & \textbf{G\_O} 
& \textbf{Q\_SC} & \textbf{Q\_PQ} & \textbf{Q\_O} 
& \textbf{G\_SC} & \textbf{G\_PQ} & \textbf{G\_O} 
& \textbf{Q\_SC} & \textbf{Q\_PQ} & \textbf{Q\_O}  \\
\midrule
Instruct-Pix2Pix~\cite{Brooks2022InstructPix2PixLT} 
& 3.335 & 6.210 & 3.234 & 4.833 & 6.992 & 4.691 & 3.296 & 6.189 & 3.219 & 4.746 & 6.913 & 4.578 \\
MagicBrush~\cite{zhang2023magicbrush} 
& 4.564 & 6.335 & 4.236 & 5.814 & 7.149 & 5.653 & 4.517 & 6.371 & 4.185 & 5.752 & 7.069 & 5.558 \\
AnyEdit~\cite{yu2024anyedit} 
& 3.122 & 5.865 & 2.919 & 3.873 & 6.754 & 3.789 & 3.053 & 5.882 & 2.854 & 3.713 & 6.730 & 3.635 \\
OmniGen~\cite{xiao2024omnigen} 
& 6.037 & 5.856 & 5.154 & 7.033 & 6.775 & 6.557 & 5.879 & 5.871 & 5.005 & 6.845 & 6.700 & 6.352 \\

Step1X-Edit
& 7.289 & 6.962 & 6.618 & 7.501 & 7.264 & 7.189 & 7.131 & 6.998 & 6.444 & 7.388 & 7.279 & 7.067 \\

\textbf{Step1X-Edit-v1.1} 
& \textbf{7.905} & \textbf{7.366} & \textbf{7.189} & \textbf{7.737} & \textbf{7.425} & \textbf{7.436} & \textbf{7.658} & \textbf{7.354} & \textbf{6.969} & \textbf{7.652} & \textbf{7.408} & \textbf{7.346} \\

\midrule
Gemini~\cite{gemini220250312} 
& 6.816 & 7.408 & 6.483 & 7.295 & 7.314 & 6.996 & 6.866 & 7.436 & 6.509 & 7.274 & 7.327 & 6.971 \\
Doubao~\cite{shi2024seededit} 
& 7.396 & 7.899 & 7.137 & 7.427 & 7.651 & 7.285 & 7.222 & 7.885 & 6.983 & 7.353 & 7.651 & 7.230 \\
GPT-4o~\cite{gpt4o20250325} 
& 7.867 & 8.097 & 7.590 & 7.905 & 7.723 & 7.752 & 7.743 & 8.133 & 7.494 & 7.847 & 7.705 & 7.692 \\
\bottomrule
\end{tabular}%
}
\caption{\textbf{Quantitative evaluation on \Bench-EN.} All metrics are reported as higher-is-better ($\uparrow$). The Intersection subset reflects the subset of prompts where all methods return valid responses with a total of 434 instances; the Full set includes all the 606 instances. G\_SC, G\_PQ, and G\_O refer to the metrics evaluated by GPT-4.1, while Q\_SC, Q\_PQ, and Q\_O refer to the metrics evaluated by Qwen2.5-VL-72B. }
\label{tab:gui_bench_en}
\vspace{-1pt}
\end{table*}

\begin{table*}[ht]
\centering
\resizebox{\textwidth}{!}{%
\begin{tabular}{l|ccc|ccc|ccc|ccc}
\toprule
\multirow{3}{*}{\textbf{Model}} 
& \multicolumn{6}{c|}{\textbf{\Bench-CN (Intersection subset)} $\uparrow$} 
& \multicolumn{6}{c}{\textbf{\Bench-CN (Full set)} $\uparrow$} \\
\cmidrule(lr){2-7} \cmidrule(lr){8-13}
& \textbf{G\_SC} & \textbf{G\_PQ} & \textbf{G\_O} 
& \textbf{Q\_SC} & \textbf{Q\_PQ} & \textbf{Q\_O} 
& \textbf{G\_SC} & \textbf{G\_PQ} & \textbf{G\_O} 
& \textbf{Q\_SC} & \textbf{Q\_PQ} & \textbf{Q\_O} \\
\midrule
Gemini~\cite{gemini220250312} 
& 5.316 & 7.571 & 5.208 & 5.658 & 7.372 & 5.566 & 5.259 & 7.600 & 5.143 & 5.622 & 7.370 & 5.525 \\
Doubao~\cite{shi2024seededit} 
& 7.228 & 7.800 & 6.915 & 7.109 & 7.687 & 7.054 & 7.168 & 7.794 & 6.837 & 7.098 & 7.676 & 7.040 \\
GPT-4o~\cite{gpt4o20250325} 
& 7.606 & 7.991 & 7.336 & 7.772 & 7.658 & 7.599 & 7.518 & 8.023 & 7.301 & 7.726 & 7.652 & 7.552 \\
\midrule
Step1X-Edit
& 7.464 & 7.076 & 6.779 & 7.527 & 7.410 & 7.259 & 7.299 & 7.142 & 6.658 & 7.490 & 7.384 & 7.212 \\
\textbf{Step1X-Edit-v1.1} 
& 7.838 & 7.401 & 7.115 & 7.636 & 7.367 & 7.327 & 7.647 & 7.398 & 6.983 & 7.532 & 7.370 & 7.240 \\

\bottomrule
\end{tabular}%
}
\vspace{10pt}
\caption{\textbf{Quantitative evaluation on \Bench-CN.} All metrics are reported as higher-is-better ($\uparrow$). The Intersection subset reflects the subset of prompts where all methods return valid responses with a total of 422 instances; the Full set includes all the 606 instances. G\_SC, G\_PQ, and G\_O refer to the metrics evaluated by GPT-4.1, while Q\_SC, Q\_PQ, and Q\_O refer to the metrics evaluated by Qwen2.5-VL-72B.
}
\label{tab:gui_bench_cn}
\vspace{-1pt}
\end{table*}

\subsection{Experimental Results}

\subsubsection{Evaluation on \textbf{\Bench}}

Based on the \textbf{\Bench}, we evaluated a diverse range of image editing algorithms, covering state-of-the-art open-source solutions such as Instruct-Pix2Pix~\cite{Brooks2022InstructPix2PixLT},  MagicBrush~\cite{zhang2023magicbrush}, AnyEdit~\cite{yu2024anyedit}, OmniGen~\cite{xiao2024omnigen}, as well as proprietary algorithms like GPT-4.1~\cite{gpt4o20250325}\footnote{The results are obtained based on ChatGPT APP in April 2025.}, doubao~\cite{shi2024seededit}\footnote{The results are obtained based on doubao APP in April 2025.},  and Gemini2 Flash~\cite{gemini220250312}\footnote{The results are obtained in April 2025.}. Following VIEScore~\cite{ku2023viescore}, we adopt three metrics: SQ (Semantic Consistency), PQ (Perceptual Quality), and O (Overall Score). SQ assesses the degree to which the edited results conform to the given editing instruction, with a score ranging from $0$ to $10$. PQ evaluates the naturalness of the image and the presence of artifacts, also using a scoring scale that ranges from $0$ to $10$. The overall score is calculated based on these evaluations. To perform the automatic evaluation for VIEScore, we adopt the state-of-art MLLM model GPT-4o\footnote{API access as of June 2025}.  Also, the evaluation based on the open-source model Qwen2.5-VL-72B~\cite{qwen2.5} is included for reproduction. To comprehensively assess model capabilities on different languages, each image in our benchmark is paired with one English (EN) and one Chinese (CN) instruction. For EN instructions (\Bench-EN), both closed and open-source models are evaluated. For CN instructions (\Bench-CN), only those models which supports Chinese prompts, i.e., the close-source systems,  are tested. During the evaluation process, we find that close-source image editing system such as GPT-4o may refuse certain instructions due to safety policies. To address this issue, we report two scores for two testing sets: (1) Intersection Subset — the subset of images whose results can be successfully returned from all the tested models, and (2) Full set — all the testing samples from \textbf{\Bench}.
For the full set of results, we will calculate the average scores only for the cases where the models successfully generate and return the target image. For each evaluated model, instances where no result image is returned due to reasons such as safety concerns will be excluded from the averaging process.

\begin{figure}
    \centering
    \includegraphics[width=0.98\linewidth]{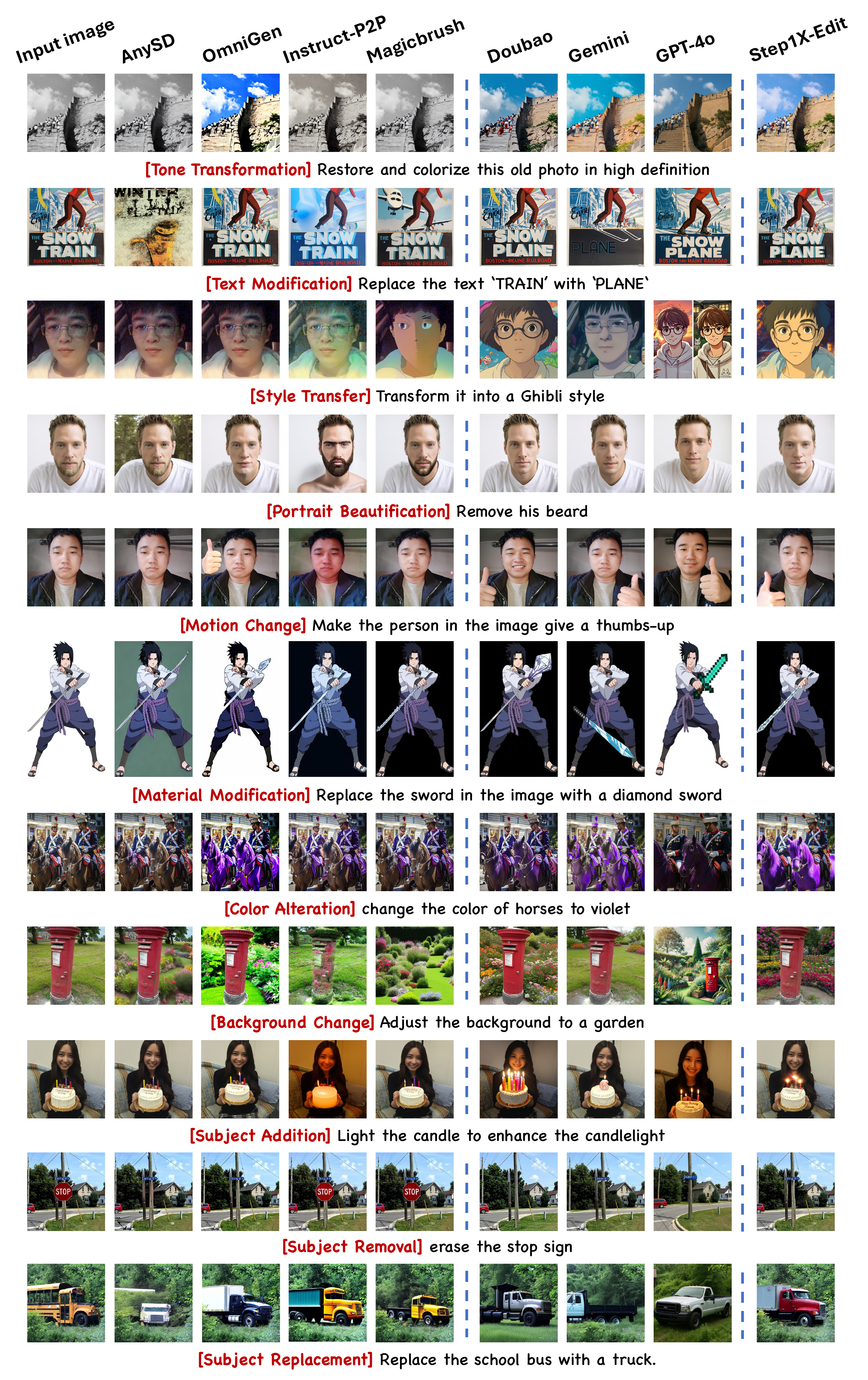}
    \caption{\textbf{A Comparative Illustration of Open-Source Approaches and Commercial systems for English Editing Instructions.}}
    \label{fig:quatlitative-en}
\end{figure}

\begin{figure}
    \centering
    \includegraphics[width=0.58\linewidth]{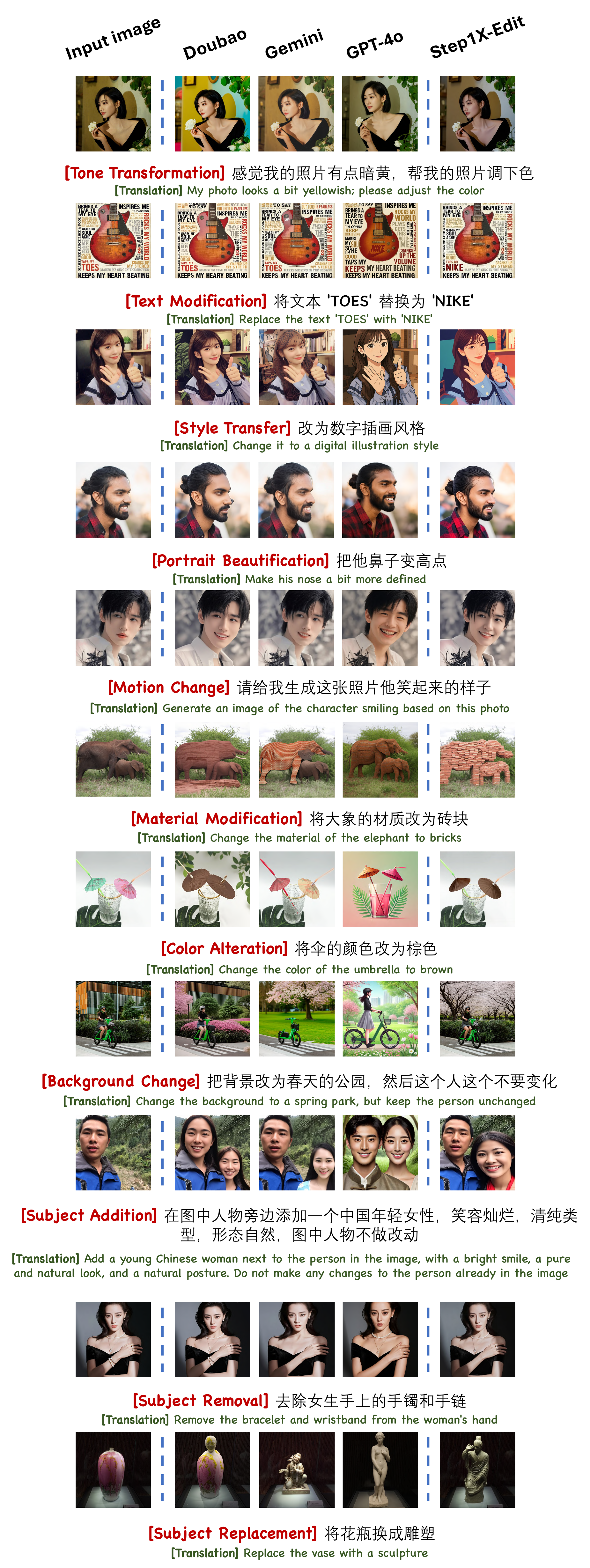}
    \caption{\textbf{A Comparative Illustration of state-of-art algorithms for Chinese Editing Instructions.}}
    \label{fig:quatlitative-cn}
\end{figure}

Fig.~\ref{fig:radar_gui_gpt} demonstrates the groundbreaking capacity of Step1X-Edit, which outperforms open-source counterparts across 11 distinct evaluation axes. When compared to closed models, it surpasses Gemini2 Flash~\cite{gemini220250312} and even beats GPT-4o~\cite{gpt4o20250325} in axes such as style change and color alteration. As detailed in Tab.~\ref{tab:gui_bench_en}, Step1X-Edit significantly outperforms the existing open-source models like OmniGen~\cite{xiao2024omnigen} and has comparable results against  the closed model like Gemini2 Flash and Doubao. Furthermore, as shown in Tab.~\ref{tab:gui_bench_cn}, Step1X-Edit demonstrates consistent performance, even surpassing Gemini2 Flash and Doubao when handling the Chinese editing instructions in \Bench-CN benchmark. These results highlight the outstanding performance of our model across all dimensions with a unified architecture, eliminating the requirement for masks during the editing process. Fig.~\ref{fig:quatlitative-en} and Fig.~\ref{fig:quatlitative-cn}  provide illustrative examples for English instructions and Chinese instructions, respectively.

\subsubsection{User Study}

To assess the subjective quality of image editing results, we conduct a comprehensive user preference study built upon the \Bench. A total of 55 participants are recruited to evaluate the outputs of four algorithms—Gemini2 Flash~\cite{gemini220250312}, Doubao\cite{shi2024seededit}, GPT-4o~\cite{gpt4o20250325}, and our method, Step1X-Edit. Each participant is presented with a series of test images and asked to rank the editing results generated by the four methods. This evaluation is performed in a blinded and subjective setting to minimize bias and ensure fairness.

Participants rate the outputs using a five-level quality scale, ranging from worst to excellent. To facilitate consistent comparison with quantitative evaluation metrics such as VIEScores, we map these qualitative ratings to numerical scores: worst = $2$, poor = $4$, fair = $6$, good = $8$, and excellent = $10$. For each editing task, we compute the mean preference score across all participants. The overall performance of each method is then summarized by averaging the scores over all editing tasks.

\begin{figure}[htbp]
    \centering

    \begin{subfigure}[b]{0.48\textwidth}
        \centering
        \includegraphics[width=\linewidth]{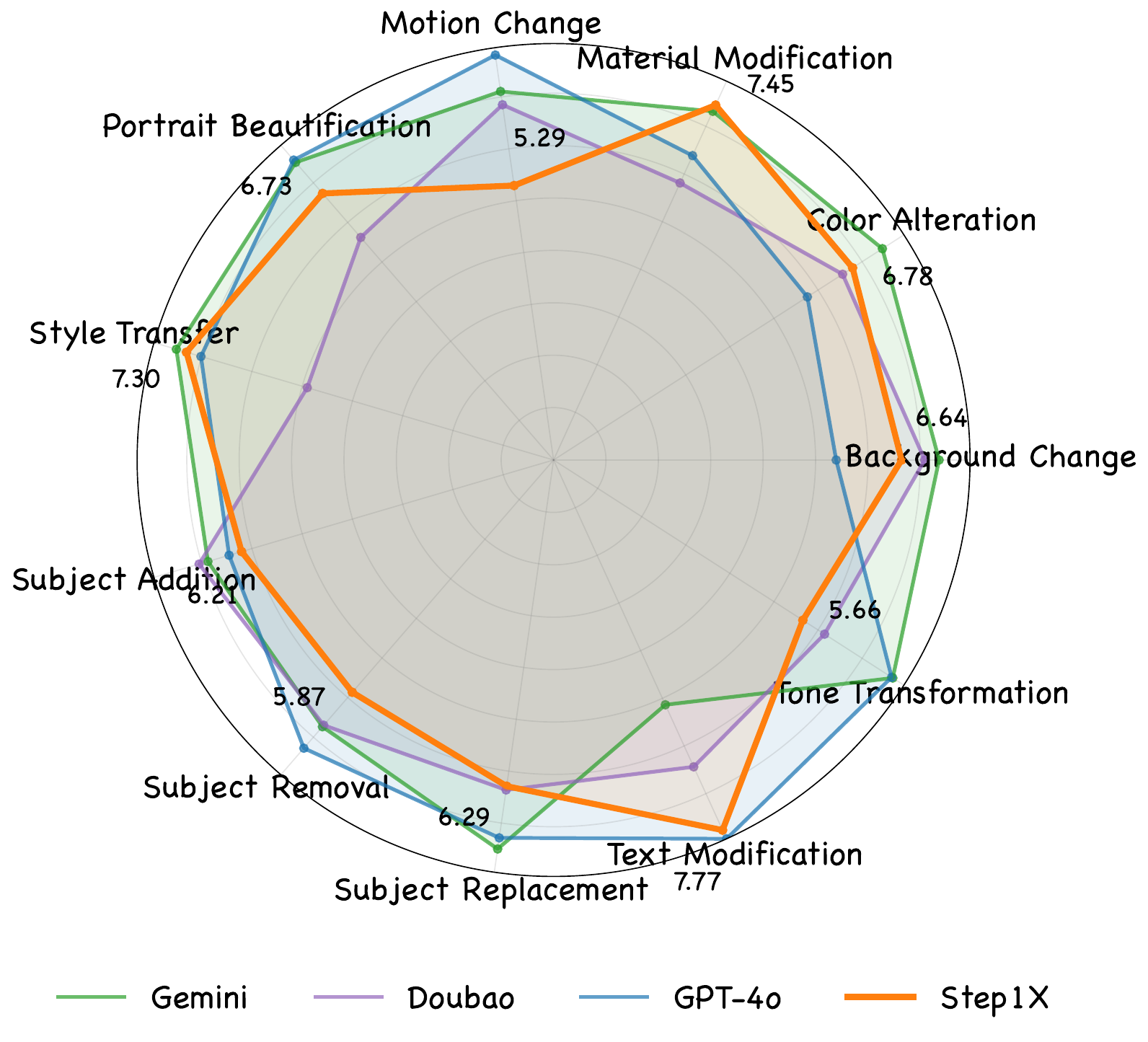}
        \caption{User Preference score in the Intersection subset.}
    \end{subfigure}
    \hfill
    \begin{subfigure}[b]{0.48\textwidth}
        \centering
        \includegraphics[width=\linewidth]{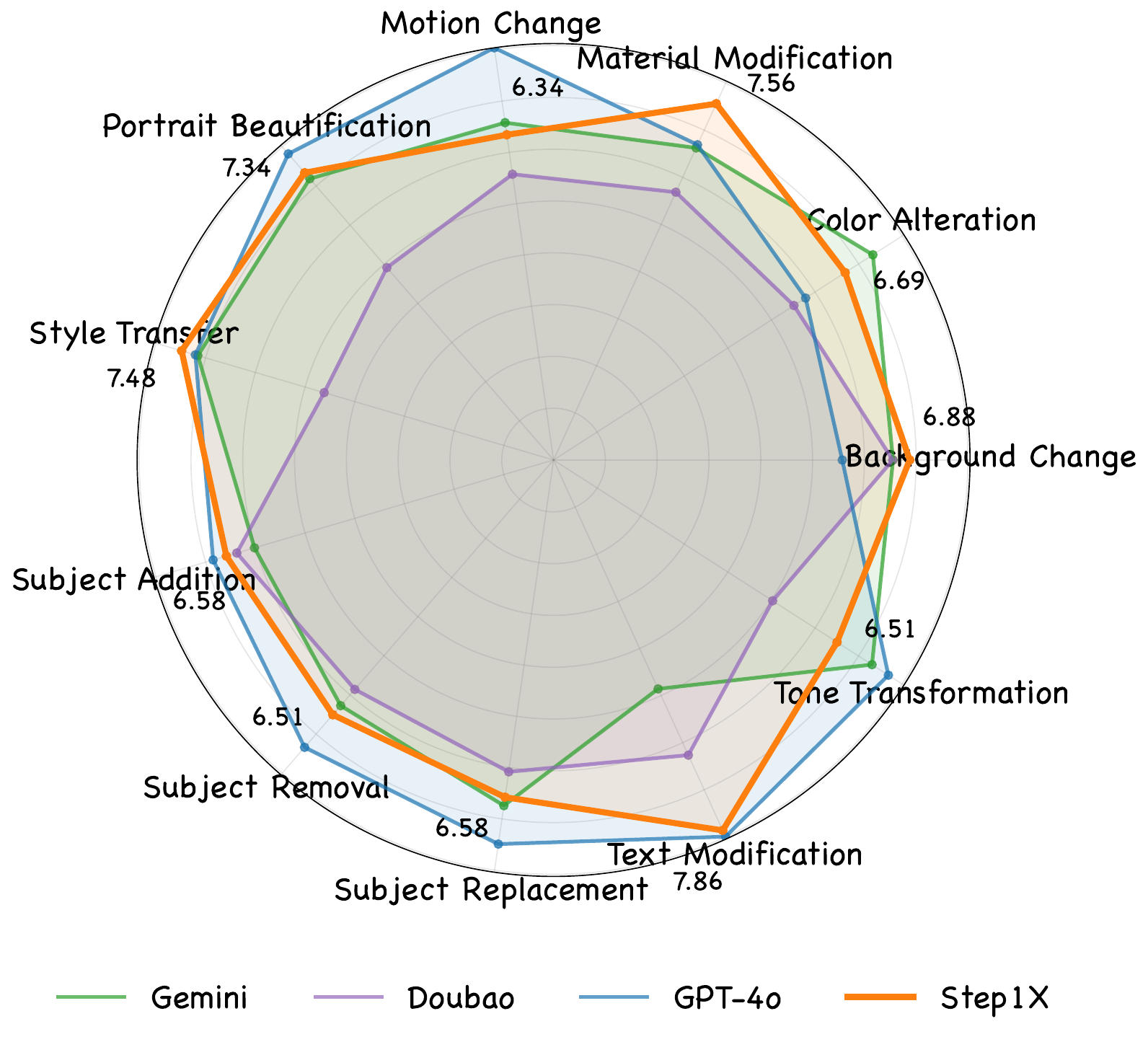}
        \caption{User Preference score in the full subset.}
    \end{subfigure}

    \caption{\textbf{User Preference score of Each Sub-task in \Bench}.}
    \label{fig:user_study_radar}
\end{figure}

The results, presented in Tab.~\ref{tab:us} and Fig.~\ref{fig:user_study_radar}, highlight the effectiveness of Step1X-Edit. Notably, our method achieves comparable subjective quality to other state-of-the-art approaches, reinforcing its capability in producing visually pleasing and user-preferred edits. It is worth noting that Gemini2 Flash achieves an astonishingly high user preference score primarily attributed to its strong identity-preserving capabilities in the testing examples.  This characteristic was more favored by the participants in the user study.

\begin{table*}[h]
\centering
\footnotesize
\begin{tabular}{l|ccc|c}
\toprule
Model & Gemini~\cite{gemini220250312} & Doubao~\cite{shi2024seededit} & GPT-4o~\cite{gpt4o20250325} & Step1X-Edit \\
\midrule
UP-IS ($\uparrow$) & 7.109 & 6.320 & 6.961 & 6.544 \\
UP-Full ($\uparrow$)& 6.603 & 5.678 & 7.134 & 6.939 \\
\bottomrule
\end{tabular}
\vspace{10pt}
\caption{\textbf{Overall user preference (UP) evaluation on \Bench.} UP-IS and UP-Full represent user preference score for Intersection subset (IS) and Full set (Full), respectively. All metrics are reported as higher-is-better ($\uparrow$).}
\label{tab:us}
\end{table*}

\section{Conclusion}
In this report, we present a new general image editing algorithm called Step1X-Edit, which will be publicly released to foster further innovation and research within the image editing community.  To train the model effectively, we propose a new data generation pipeline which can generate large-scale high-quality image editing triples, each consisting of a reference image, an editing instruction, and a corresponding target image. Based on the collected dataset, we train our Step1X-Edit model by seamlessly integrating powerful Multimedia Large Language Model  with a diffusion-based image decoder. According to the evaluations on our collected \Bench, our proposed algorithm outperforms the existing open-source image editing algorithms with a substantial margin. 

\section*{Contributors and Acknowledgments}

We designate core contributors as those who have been involved in the development of Step1X-Edit throughout its entire process, while contributors are those who worked on the early versions or contributed part-time.

\textbf{Core Contributors:} 
Shiyu Liu, Yucheng Han, Peng Xing, Fukun Yin,  Rui Wang, Wei Cheng, Jiaqi Liao, Yingming Wang, Xianfang Zeng, Gang Yu.

\textbf{Contributors:} Honghao Fu, Ruoyu Wang, Yongliang Wu, Tianyu Wang, Haozhen Sun, Wen Sun, Bizhu Huang, Mei Chen, Kang An, Shuli Gao, Wei Ji, Tianhao You, Chunrui Han, Guopeng Li, Yuang Peng, Quan Sun, Jingwei Wu, Yan Cai, Zheng Ge, Ranchen Ming, Lei Xia, Yibo Zhu, Binxing Jiao, Xiangyu Zhang, Daxin Jiang.

\textbf{Corresponding Authors:} Xianfang Zeng(zengxianfang@stepfun.com),   Gang Yu (yugang@stepfu\\n.com), Daxin Jiang (djiang@stepfun.com).

{
    \small
    \bibliographystyle{plain}
    \bibliography{main}
}

\end{document}